\documentclass[10pt,  letterpaper]{IEEEtran}
\usepackage{amsmath,amssymb,latexsym,float,epsfig,subfigure, textcomp,hyperref}
\newcommand{\dg}{\ensuremath{^\circ}}

\newcommand{\tp}{\mathbf{tp}}
\newcommand{\mm}{\mathbf{m}}
\begin{document}

\author{Maxime Gariel\thanks{Ph.D Candidate, Georgia Institute of Technology, 270 Ferst Drive, Atlanta, GA, 30332} \and Ashok N. Srivastava \thanks{Principal Investigator, Integrated Vehicle Health Management; Group Lead, Intelligent Data Understanding, NASA Ames Research Center, Moffett Field, CA} \and Eric Feron \thanks{Dutton/Ducoffe professor of Aerospace Engineering, Georgia Institute of Technology, 270 Ferst Drive, Atlanta, GA, 30332.}}

\title{Trajectory Clustering and an Application to Airspace Monitoring}


\maketitle

\begin{abstract}
 This paper presents a framework aimed at monitoring the behavior of aircraft in a given airspace. Nominal trajectories are determined and learned using data driven methods. Standard procedures are used by air traffic controllers (ATC) to guide aircraft, ensure the safety of the airspace, and to maximize the runway occupancy. Even though standard procedures are used by ATC, the control of the aircraft remains with the pilots, leading to a large variability in the flight patterns observed. 
Two methods to identify typical operations and their variability from recorded radar tracks are presented. This knowledge base is then used to monitor the conformance of current operations against operations previously identified as standard. A tool called AirTrajectoryMiner is presented, aiming at monitoring the instantaneous health of the airspace, in real time. The airspace is ``healthy'' when all aircraft are flying according to the nominal procedures. A measure of complexity is introduced, measuring the conformance of current flight to nominal flight patterns. When an aircraft does not conform, the complexity increases as more attention from ATC is required to ensure a safe separation between aircraft. 
\end{abstract}

\section*{Introduction}
To address the challenges of increase in air traffic volume, new technologies and procedures are being developped in the context of NextGen~\cite{nextGen} in the US and SESAR~\cite{SESARdeliverable5} in Europe. Automation is a key element, necessary to achieve the goals set by those programs. New procedures involving more accurate navigation are predicted to increase the capacity of the airspace. Analyzing trajectory records is a key element to assess the performances and the accuracy of new concepts of operations. Automated tools are needed to process the large amount of daily flights and corresponding records. This work presents two methods to cluster trajectories and identify flights that followed identical air routes. The first method is based on the identification of waypoints in the trajectories, and the second method is based on a principal components analysis of resampled trajectories.  Operations in the terminal area are managed by Air Traffic Controllers (ATC) and are not part of the flight plans. It was therefore decided not to use any flight plan knowledge or aircraft intent other than the destination airport. 
Then using the knowledge gathered from the clustering methods, we propose a real time airspace monitoring tool that evaluates the conformance of current flight to pre-identified nominal trajectories. A measure of airspace complexity based on this conformance is also proposed with the tool. The overall method developed in this paper is neither location nor data specific and can easily be adapted to other data sets since unsupervised methods are used, and the data is not labeled.  This paper considers radar tracks in a terminal radar approach control (TRACON). However, the underlying principles may also be used for other applications, such as a GPS-equiped fleet of trucks. Since this paper deals with different problems such as trajectory clustering, airspace monitoring and airspace complexity, the literature review is spread along the paper at the begining of the corresponding section. The remainder of this paper is organized as follows: The first section presents the data set used for the study. The second section presents the trajectory clustering methods, and finally, before the concluding remarks, the third section introduces AirTrajectoryMiner, the airspace monitoring tool that detects in real time the aircraft that do not comply to nominal procedures.

\section{Available Data}
The available data~\footnote{The complete dataset is available for download ~\url{http://dashlink.arc.nasa.gov}} consists of records of flight tracks over the San Francisco bay area, for the first 3 months of 2006. The records cover the Northern California TRACON (NCT), that is, a cylinder of radius 80km and height 6,000m centered at Oakland international airport.  The NCT contains 3 main airports \texttwelveudash~Oakland, San Francisco (SFO) and San Jose International airports \texttwelveudash~ as well as many smaller airports. The NCT is the fourth busiest terminal area in the US~\cite{FAAfactBook2006} with an average of 133,000 flight instrument operations per month in 2006. The data, made of the position and speed of aircraft, is organized by flight and also contains metadata for each flight that include: type of operation (departure/arrival), origin and destination airports, aircraft type (business, jet, helicopter, other, etc), date and time of beginning of record, duration of the record, etc. 

\begin{figure}
\centering 
\includegraphics[width = 0.25\textwidth]{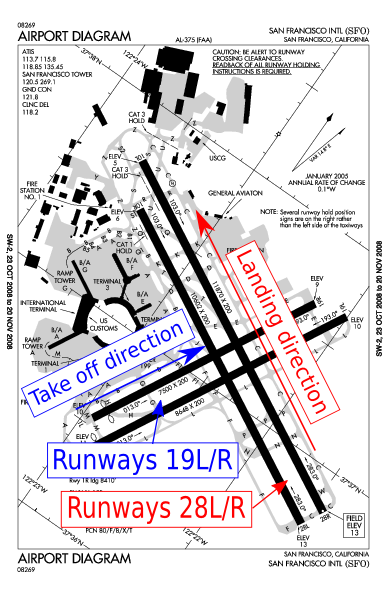}\caption{San Francisco airport diagram with take off and landing direction in the west configuration }\label{fig:sfoDiag}
\end{figure}

\begin{figure}
\centering
\includegraphics[width = 0.45\textwidth ]{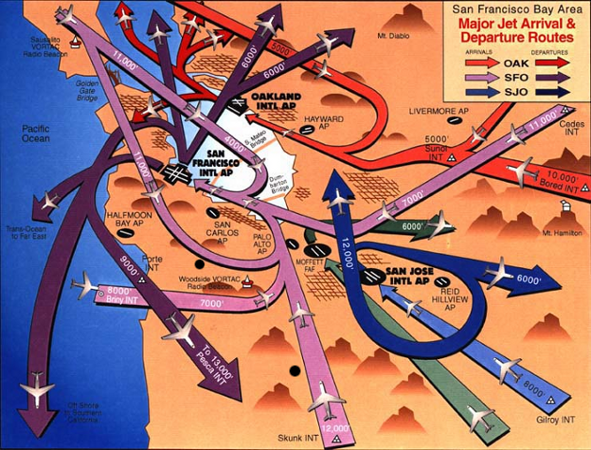}\caption{NCT standard traffic patterns, west configuration, image courtesy of Federal Aviation Administration}\label{fig:NCTtrafficPatterns}
\end{figure}

Using the available metadata, visual flight rules (VFR) traffic is discarded, since it is more unpredictable and does not follow the same rules as instrument flight rules (IFR) traffic. The metadata is used to sort trajectories by airport and operation type, i.e. take off or landing. After a visual analysis of the flight patterns for the different airports, it was decided to focus the study on the landings at SFO. It is the busiest airport in the NCT and the arrival tracks present the most interesting patterns by their numbers and variety. The most frequent configuration is the ``West'' configuration, where aircraft land on runways 28L/R and take off from runways 19L/R. A diagram of SFO is presented in Figure \ref{fig:sfoDiag}, and Figure \ref{fig:NCTtrafficPatterns} depicts the NCT traffic patterns typically used in the west configuration. 


In this paper, the axes are set by the radar, located at ${(0,0,0)}$. The $x$ and $y$ axis define the horizontal plan and $z$ the vertical direction, positive going upward. To each recorded flight, corresponds an aircraft $i$ and a trajectory ${T_i, i=1\ldots n}$, where $n$ is the total number of trajectories of interest in the dataset. Each trajectory $T_i$ is a $m_i \times 4$ matrix, and the line $T_i^l$ of $T_i$ is the $l^{\text{th}}$ radar echo, given by ${T_i^l = (x_{i}^l,y_{i}^l,z_{i}^l,t_i^l)}$, where ${(x_{i}^l, y_{i}^l, z_{i}^l)}$ is the 3 dimensional coordinates of aircraft $i$ at time $t_i^l$. The trajectories have different numbers of points $m_i$, varying from 10 to about 550 points, depending on the duration of the trajectory. Trajectories with a few datapoints correspond to trajectories on the boundaries of the dataset, or to short flights from San Jose International Airport or Oakland Internation Airport to SFO. The interval between points is between 4 and 5 seconds and is given by the rotational speed of the radar (most likely 4.8 sec). The time stamp $t_i^l$ is rounded to the nearest second.


\section{Trajectory Clustering}
In this section, two trajectory clustering methods are presented. After a review of existing trajectory clustering methods, a technique based on trajectories' ``waypoints'' is presented. Then, a technique based on a principal component analysis of resampled and augmented trajectories is introduced.

\subsection{Literature review}

The use of positioning devices such as GPS and the collection of data has increased over the past 15 years leading to an increasing number of tracking applications. An objective of tracking is to discover common patterns on the one hand, and detect outliers on the other hand. 

Piciareli et al.~\cite{piciarelli2005trajectory} presented an on-line trajectory clustering method for real time video surveillance. Moving objects such as pedestrians are identified in video frames and their trajectories are compared against existing cluster representatives, that is, an average of all the trajectories in the cluster. The match between a trajectory and a cluster is determined using the mean of the normalized distances of every trajectory point to the nearest point of the cluster representative. If a match is found, the cluster representative is updated. If not, a new cluster is created. In this approach, the cluster representatives evolve with time. This clustering method was used by Dahlbom and Niklasson for coastal surveillance but failed to provide satisfactory results when dealing with real data sets~\cite{dahlbom2007trajectory} such as ship trajectories. 

Lee et al.~\cite{lee2007trajectory} presented a partition-and-group framework for trajectory clustering. Trajectories are partitionned in subtrajectories. Subtrajectories are represented by line segments and grouped using a distance function. The distance function incorporates three components that measure the perpendicular distance, the parallel distance and the angular distance between the line segments. The clustering algorithm is density based, i.e clusters are created where the density of points is the highest. The formulation is powerful but the results are presented on very noisy data where it is difficult to visually cluster the trajectories.There exists no well-defined measure to assess the results of the clustering method. Based on the same distance measure, Lee et al.~\cite{lee2008trajectory} present a trajectory outlier detection procedure. The results are presented on the same noisy datasets and therefore difficult to evaluate visually.   

Vlachos et al. used similarity functions based on the longest common subsequence (LCS) to discover similar multidimensional trajectories~\cite{vlachos2002discovering}. Their LCS based clustering method appears to be more efficient than Euclidean distance based measures and dynamic time warping distance functions, especially in the presence of noise. Dynamic time warping allows the tool to measure distance between sequences which may vary in time or speed. 

Eckstein proposed an automated flight track taxonomy~\cite{eckstein09taxonomy} . The trajectories are first resampled, then clustered using $k$-means on a reduced order model. The model reduction is the truncation of a proper orthogonal decomposition (POD), also called principal components analysis. The trajectories are clustered using only the first two modes of the decomposition, as they capture 95\% of the fluctuations of the dataset used.

\subsection{Waypoint based trajectory clustering}
This section presents a novel algorithm for aircraft trajectory clustering. This algorithm takes advantage of aircraft trajectory properties: aircraft usually fly straight, with a limited number of turns. This method arises from the current  instrument flight rules procedures. When approaching an airport, aircraft usually follow published procedures made of a sequence of waypoints. A waypoint is characterized by its GPS coordinates and, sometimes, an altitude indication. The planar localization of a waypoint is very accurate but its vertical component often looks like ``at or above ---ft''. Vertical clearances are delivered by ATC and trajectories vertical profiles are then at the discretion of the pilots. Therefore, this method focuses on the 2D coordinates of the waypoints in the $(x,y)$ plan. This method is an efficient way to determine the compliance of flown trajectories with published procedures. Nevertheless, published procedures cannot be used because of the limited number of waypoints or reporting points located in the TRACON. In Section \ref{sec:airspaceMonitoring}, we further show this by comparing the results of the trajectory clustering with the published waypoints.\\

\begin{figure}
 \centering
\includegraphics[width=0.45\textwidth]{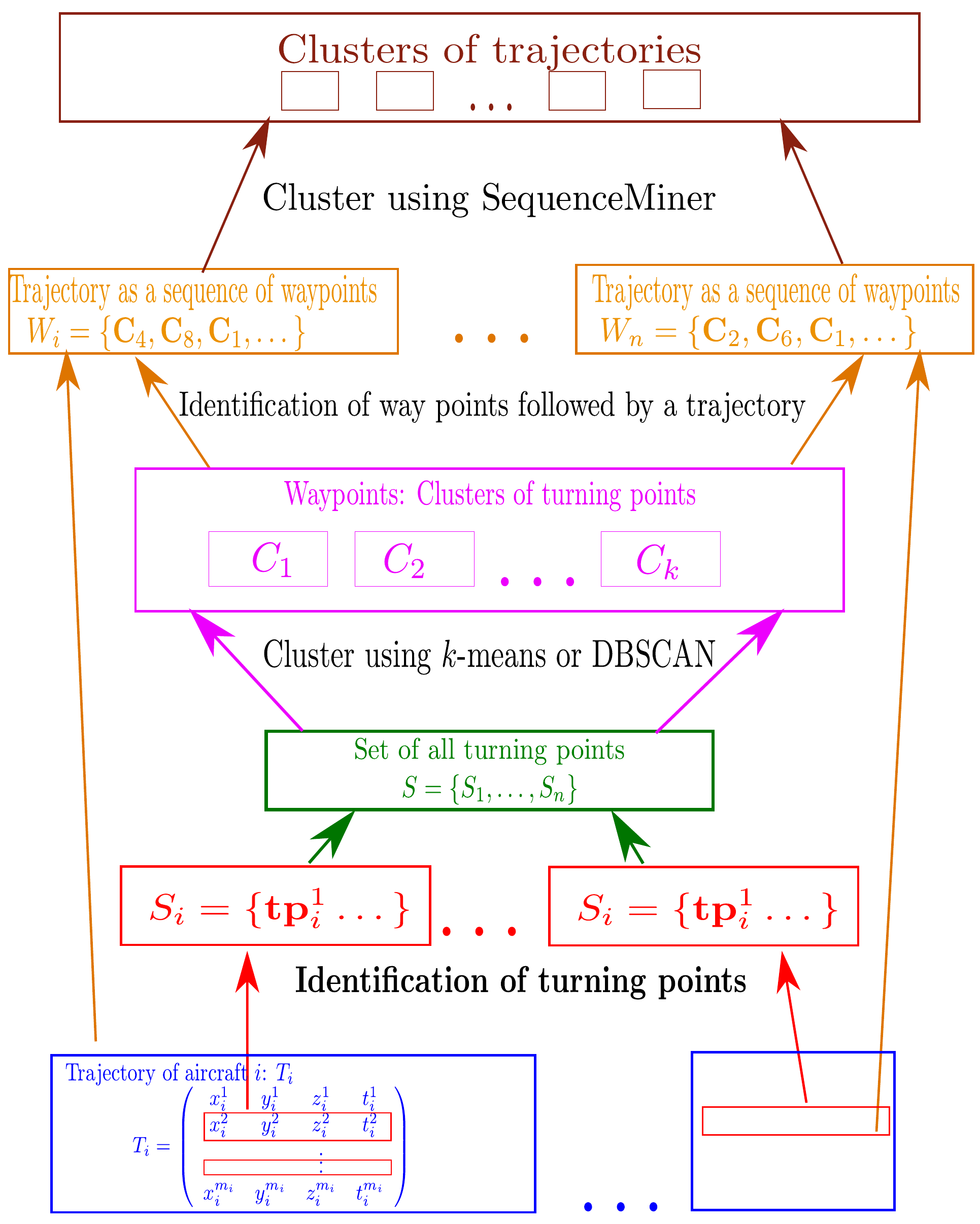}
\caption{Waypoint clustering method}\label{fig:trajectoryClusteringWayPointsDiagram}
\end{figure}

The objective is to identify and group the turning points into ``waypoints''. A turning point is a point in the trajectory where the aircraft changes heading. Then trajectories are represented by a sequence of waypoints.  Finally, trajectories are clustered using the Longuest Common Subsequence (LCS). The algorithm proceeds using the following steps and it is summarized in Figure \ref{fig:trajectoryClusteringWayPointsDiagram}:

\begin{enumerate}
\item Identify the location of the turning points of each trajectory. 
\item Cluster of the set of all the turning points of all the trajectories.  This clustering task is done using $k$-means~\cite{mackay2003information, HastieTibshirani} or DBSCAN~\cite{ester1996density} (Density-Based Spatial Clustering of Applications with Noise). Section \ref{sec:algosClustering} gives an overview of those algorithms. This clustering provides a finite number of waypoints where it has been determined that aircraft usually turn. 
\item Represent each trajectory by its sequence of turning points. 
\item Cluster the sequences of waypoints using the SequenceMiner algorithm~\cite{budalakoti2005anomaly, budalakoti2009anomaly}. SequenceMiner provides us with a representative  trajectory for each cluster. 
\end{enumerate}

\subsubsection{Turning Points Identification}
The first step is to extract the location of the turning points of each trajectory. To simplify the notations, the aircraft index $i$ is omitted in the following equations. The heading $\Psi^l$ of an aircraft at time $t^l$ can be estimated by ${\psi^l = \arctan\frac{y^{l+1}-y^{l-1}}{x^{l+1}-x^{l-1}}}$, at each point of the trajectory,  $l$, $l=2\ldots m-1$, where $m$ is the total number of points. 
Since the trajectory is a bit noisy, a low pass filter is applied:
\begin{align}\label{eq:headingFiltering}
 \tilde \psi^1 &= \psi^1\\
 \tilde \psi^l &= \alpha \psi^{l} + (1-\alpha)\tilde \psi^{l-1}, \quad l=2\ldots m-1,
\end{align}
where $\alpha$ is a constant for the filter. On this data, setting $\alpha = 0.4$ provided good noise filtering results and not too much delay. A turning point $\tp$ is identified when the heading difference between two consecutive values of the heading exceed a threshold: $|\tilde \Psi^l - \tilde \Psi^{l-1}| > \Psi_c$. The threshold was chosen relatively small in order to capture small heading changes but not too small not to capture meaningless heading changes variations: ${\Psi_c = 0.025 \text{rad} = 1.43\dg}$. This value was set experimentally. The results are not very sensitive to a small change in $\Psi_c$. The number of turning points is trimmed to avoid long sequences when aircraft are executing large turns: if two consecutive turning points are determined, the fist one only is kept; if three, only the midlle one, etc. 

\begin{figure}[ht]
\centering
\includegraphics[width = 0.45\textwidth]{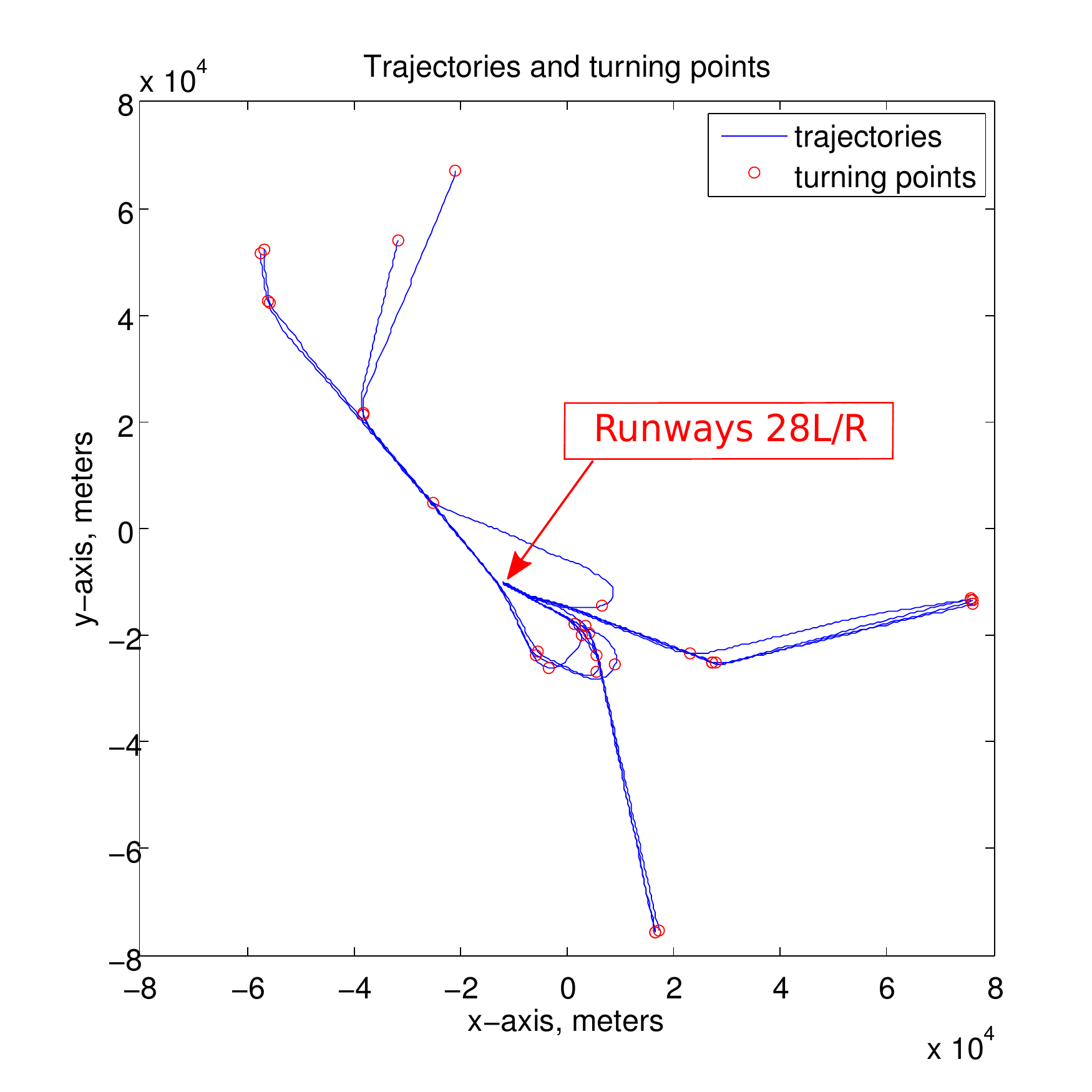}
\caption{Trajectories and identified turning points}\label{trajectorySequence}
\end{figure}

The trajectory of aircraft $i$ is now represented as a sequence of turning points $S_i$ : 
\begin{equation*}
S_{i} =  \{\tp_{i}^1 \ldots \tp_{i}^s\}, 
\end{equation*}
where  $\tp_{i}^s$ is the 3D coordinate of a turning point. The first point of the trajectory is labeled as a turning point.
Figure \ref{trajectorySequence} presents a sample of 11 trajectories and the points identified as turning points. 

Denote by $S$ the set of all the turning points for all the trajectories: $S = \{S_{1} \ldots S_{n}\}.$

The second step is to cluster the set $S$ of turning points. The following section introduces the two clustering algorithms used in this paper: $k$-means~\cite{macqueen1966some} and DBSCAN~\cite{ester1996density}.

\subsubsection{Algorithms overview: $k$-means and DBSCAN}\label{sec:algosClustering}
\begin{figure}[ht]
\centering
\includegraphics[width = 0.45\textwidth]{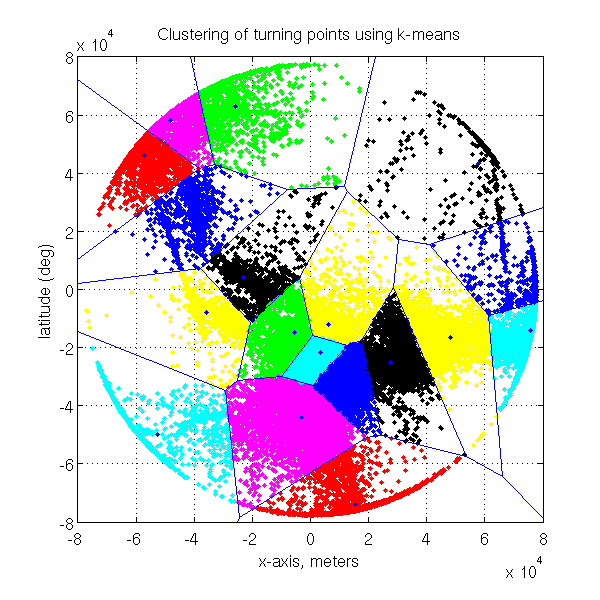}
\caption{Clusters of turnings points and corresponding Voronoi diagram}\label{fig:allTurningPointsVoronoi}
\end{figure}

\paragraph{Overview of $k$-means~\cite{macqueen1966some}}
This paragraph presents a brief overview of the $k$-means algorithm. For more details, the reader is refered to \cite{HastieTibshirani}.
Given a set ${S = (\tp_1, \dots \tp_{|S|})}$ of $|S|$ observations (turning points in our case), where each observation is a $d$-dimensional real vector, then $k$-means clustering aims at partitioning the $|S|$ observations into $k$ sets, or clusters, ${(k < |S|), C = \{C_1, C_2, \ldots, C_k}\}$ so as to minimize the within-cluster sum of squares:
\begin{equation}\label{eq:WCSS}
\underset{\mathbf{C}}{ \arg\min} \sum_{i=1}^{k} \sum_{\mathbf \tp_j \in C_i} \big\| \mathbf \tp_j - \mm_i \big\|^2 
\end{equation}
where $\mm_i$ is the mean of $C_i$. The mean $\mm_i$ of a cluster is called centroid and is the center of mass of all the elements in the cluster. The number $k$ of clusters is the only input required from the user.\\ 
Starting with an initial set of $k$ centers $\mm_1^{(1)},\ldots,\mm_k^{(1)}$, which may be specified randomly or by some heuristic, the algorithm proceeds by alternating between two steps, also known as Lloyd Algorithm~\cite{lloyd1982least}:

\textbf{Assignment step}: Assign each observation to the cluster with the closest mean, that is partition the observations according to the Voronoi diagram generated by the centroids of the clusters. Figure \ref{fig:allTurningPointsVoronoi} presents the results of $k$-means clustering and the corresponding Voronoi diagram. 
\begin{equation}
 \begin{split}
C_i^{(t)} &= \{ \mathbf \tp_j : \big\| \mathbf \tp_j - \mathbf m^{(t)}_i \big\| \leq \big\| \mathbf \tp_j - \mathbf m^{(t)}_{i^*} \big\|, \\
& \text{ for all }i^*=1,\ldots,k \}    
 \end{split}
\end{equation}

\textbf{Update step}: Calculate the new means to be the centroid of the observations in the cluster.
\begin{equation}
\mathbf m^{(t+1)}_i = \frac{1}{|C^{(t)}_i|} \sum_{\mathbf \tp_j \in C^{(t)}_i}  \tp_j 
\end{equation}

The algorithm is deemed to have converged when the assignments no longer change. Since it is a heuristic algorithm, there is no guarantee that it will converge to the global optimum, and the result may depend on the initial clusters. Since the algorithm is usually very fast, it is common to run it multiple times with different starting conditions and keep the run that resulted in the minimum value for equation \ref{eq:WCSS}. 

\paragraph{Overview of DBSCAN}
This paragraph presents a brief overview of the DBSCAN algorithm. For more details, the reader is refered to \cite{han2006data}.
DBSCAN~~\cite{ester1996density} stands for Density-Based Spatial Clustering of Applications with Noise. DBSCAN clusters points that are close together (in an $\epsilon$ neighborhood), and surrounded by sufficiently many points. DBSCAN requires two parameters: a real,$\epsilon$, and the minimum number of points, $MinPts$, required to form a cluster.  The  $\epsilon$-neighborhood of a point $p$ consists of all the points $q$ s.t $dist(p,q) \leq \epsilon$. 
If the $\epsilon$-neighborhood of a point $p$ contains more than $MinPts$, a new cluster is started, with $p$ as a core object. DBSCAN then iteratively collects directly density-reachable objects from these core objects. An object $q$ is said to be directly density-reachable from an object $p$ if $q$ is in the $\epsilon$-neighborhood of $p$ and $p$ is a core object. 

If a core object $q$ of a cluster $C_q$ is added a cluster $C_p$, $C_p$ and $C_q$ are merged.  When no point can be added to any cluster, the process terminates.

\subsubsection{Turning Points Clustering; creation of waypoints}
To determine the waypoints, the turning points are clustered: a waypoint is defined as the planar $(x,y)$ coordinates of a cluster of turning points. The idea is to create a waypoint where it has been determined that many aircraft turned. Depending on the number and on the density of available turning points , two different algorithms are used. When the spatial distribution of turning points is sparse, $k$-means is used, and when the distribution of turning points is dense, DBSCAN is used.

\begin{figure}[ht]
\centering
\includegraphics[width = 0.45\textwidth, height = 0.43\textwidth]{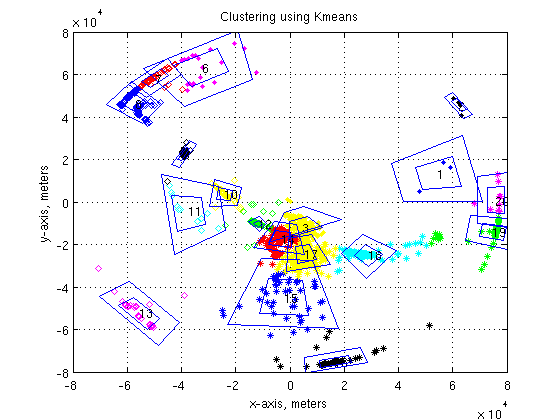}
\caption{Result of the clustering of the turning points for one day}\label{wayPointsClusters}
\end{figure}

\paragraph{Case when the data is sparse}
When the number of turning points is small, a density based clustering algorithm would provide poor results, identifying most of the points as outliers. Therefore, a distance-based algorithm is used so all the turning points available are used. A waypoint is created for each cluster produced by $k$-means. Using cylindrical coordinates, the coordinates of the center of a waypoint are given by $(r_m\, theta_m)$. The center is the center of mass of all the points in the cluster. The coordinates of the corners of the waypoints are given by ${\{(r_m + 2std_r, \theta_m + 2std_\theta),(r_m - 2std_r, \theta_m + 2std_\theta), (r_m - 2std_r, \theta_m - 2std_\theta), (r_m + 2std_r, \theta_m - 2std_\theta)\}}$, where $std_r$ and $std_\theta$ are the standard deviation of the radial coordinates and angular coordinates  of the points in the cluster, respectively.
Figure \ref{wayPointsClusters} presents the outcome of clustering the waypoints for one day of trajectories. Each cluster is represented using a different color/shape combination. The waypoints and represented by pairs of nested polygons on the figure. The inside polygon corresponds to $(r_m \pm std_r, \theta_m \pm std_\theta)$ and the outside one to $(r_m \pm 2std_r, \theta_m \pm 2std_\theta)$. The number is the label of the cluster.

\paragraph{Case when the data is dense}\label{sec:DBSCAN}
When the the number of turning points is large, a large share of the airspace is covered with turning points. A distance based algorithm such as  $k$-means provides meaningless clusters for our application. Figure \ref{fig:allTurningPointsVoronoi} shows the clusters provided by $k$-means and the corresponding Voronoi diagram for the turnings point of almost 3 month of data ( 30,000 trajectories).

\begin{figure}[ht]
\centering
\includegraphics[width = 0.45\textwidth]{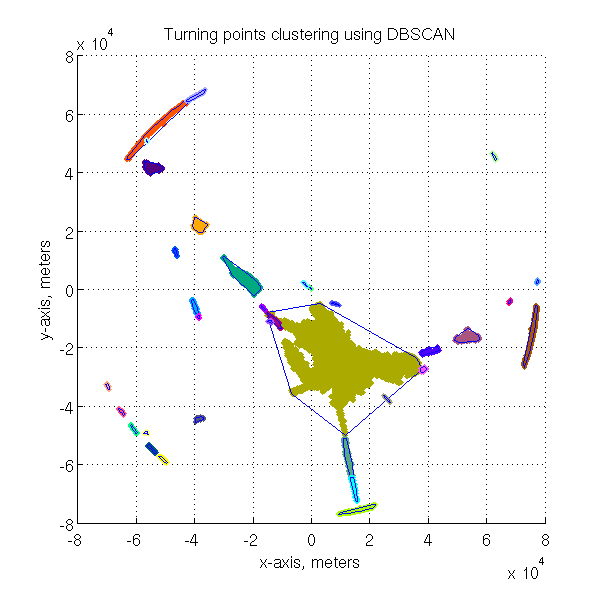}
\caption{Clusters of turning points using DBSCAN. Outliers are not displayed}\label{fig:clustersTurningPointsDBSCAN}
\end{figure}

To overcome this issue, the turning points were clustered using DBSCAN. DBSCAN is particularly efficient to cluster data in the presence of noise. Waypoints are created using the convex hull of the clusters resulting from DBSCAN. Figure \ref{fig:clustersTurningPointsDBSCAN} shows the result of the clustering of the turning points using DBSCAN. The blue polygons represent the waypoints. All the points identified as outliers, i.e not associated with any waypoint, are not depicted. The parameters used were $\epsilon = 350$m and $minPts = 10$. The main issue with DBSCAN is its execution time since its complexity is in $O(n \log n)$. Here, the number of turning points to cluster is $n = 118,179$ for 30,000 trajectories.  

\subsubsection{Converting a trajectory into a sequence of waypoints}
The waypoints have been discovered using the turning points of the trajectories. Nevertheless, some trajectories might go over waypoints without actually turning. To identify the sequence of waypoints followed by a trajectory, the following procedure is used for each trajectory: start with an empty sequence of waypoints, and given the set of all waypoints, run the trajectory along its original direction. If one of the points is located over a waypoint, the waypoint is added to the sequence. Each trajectory is now represented as an ordered sequence of waypoints, where the number of waypoints is finite. The next step is to cluster the trajectories determining the longest common subsequence (LCS) of waypoints.

\subsubsection{Longest Common Subsequence determination}
The sequences of waypoints are clustered using the longest common subsequence. The LCS problem is to find the longest subsequence common to all sequences in a set of sequences. SequenceMiner~\cite{budalakoti2005anomaly, budalakoti2009anomaly} is an algorithm that identifies the LCS and generates clusters of sequences. This method allows to cluster of sequences that do not contain the same number of elements. 
When sequences have only a small number of points, say fewer than 3, this clustering method does not work well. Therefore, only the sequences containing more than 4 waypoints are kept. The total number of waypoints being small, it is preferable to focus on the waypoints at the begining of the trajectory: since most aircraft do a final turn to get aligned with the runway, this turning point does not bring much information about the trajectory. Therefore, if the last turning point is in the large brown cluster (Figure \ref{fig:clustersTurningPointsDBSCAN}), it is removed from the sequence. 

\begin{figure}
\centering
\includegraphics[width = 0.45\textwidth]{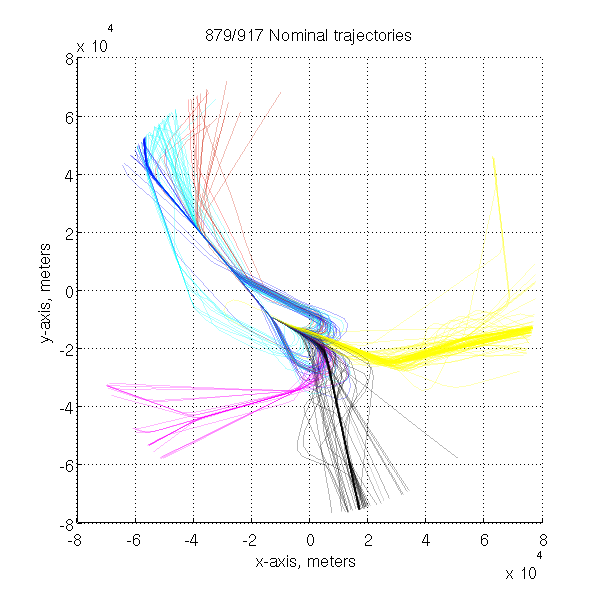}
\caption{Results of trajectory clustering for the landings of one day at SFO}\label{fig:clustersSFOland}
\end{figure}

\begin{figure}
\centering
\includegraphics[width = 0.45\textwidth]{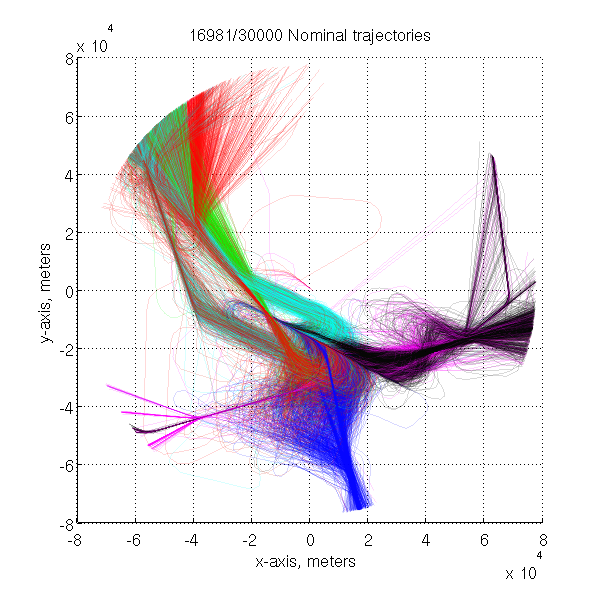}
\caption{Results of trajectory clustering for 30,000 trajectories}\label{fig:trajectoryClustersWaypointsDBSCAN}
\end{figure}

Figure \ref{fig:clustersSFOland} presents the results of the clustering process using $k$-means and LCS on a low number of trajectories. The dataset used is the tracks of all the aircraft landing at San Francisco (SFO) airport on February 10, 2006. Only the trajectories of that day were used to determine the waypoints. Each cluster is represented by a color. The algorithm identifies the main flows but a few trajectories seem not to belong to the expected cluster. The quality of the results is subjective and can only be visually assessed. 
 Figure \ref{fig:trajectoryClustersWaypointsDBSCAN} presents the results for an initial set of 30,000 trajectories, using DBSCAN and LCS. Here, the denomination ``Nominal'' qualifies the trajectories containing more than 4 waypoints. The colors corresponds to the clusters. They differ on figures \ref{fig:clustersSFOland} and \ref{fig:trajectoryClustersWaypointsDBSCAN} because the indexing of clusters is random and depends on the order of the data in the dataset.

e clus
Overall, this method presents good clustering results. One of the main drawbacks of this method is that it only keeps the trajectories going over the waypoints. For instance, consider two parallel trajectories: one going over the waypoints and the other one slightly off. The latter will be considered as an outlier eventhough it is very similar to the first trajectory, resulting in excluding many trajectories. In addition, trajectories containing large rerouting periods will belong to the clusters as long as they pass over waypoints.

\subsection{Trajectory-based clustering via component analysis}
\begin{figure}
 \centering
\includegraphics[width=0.45\textwidth]{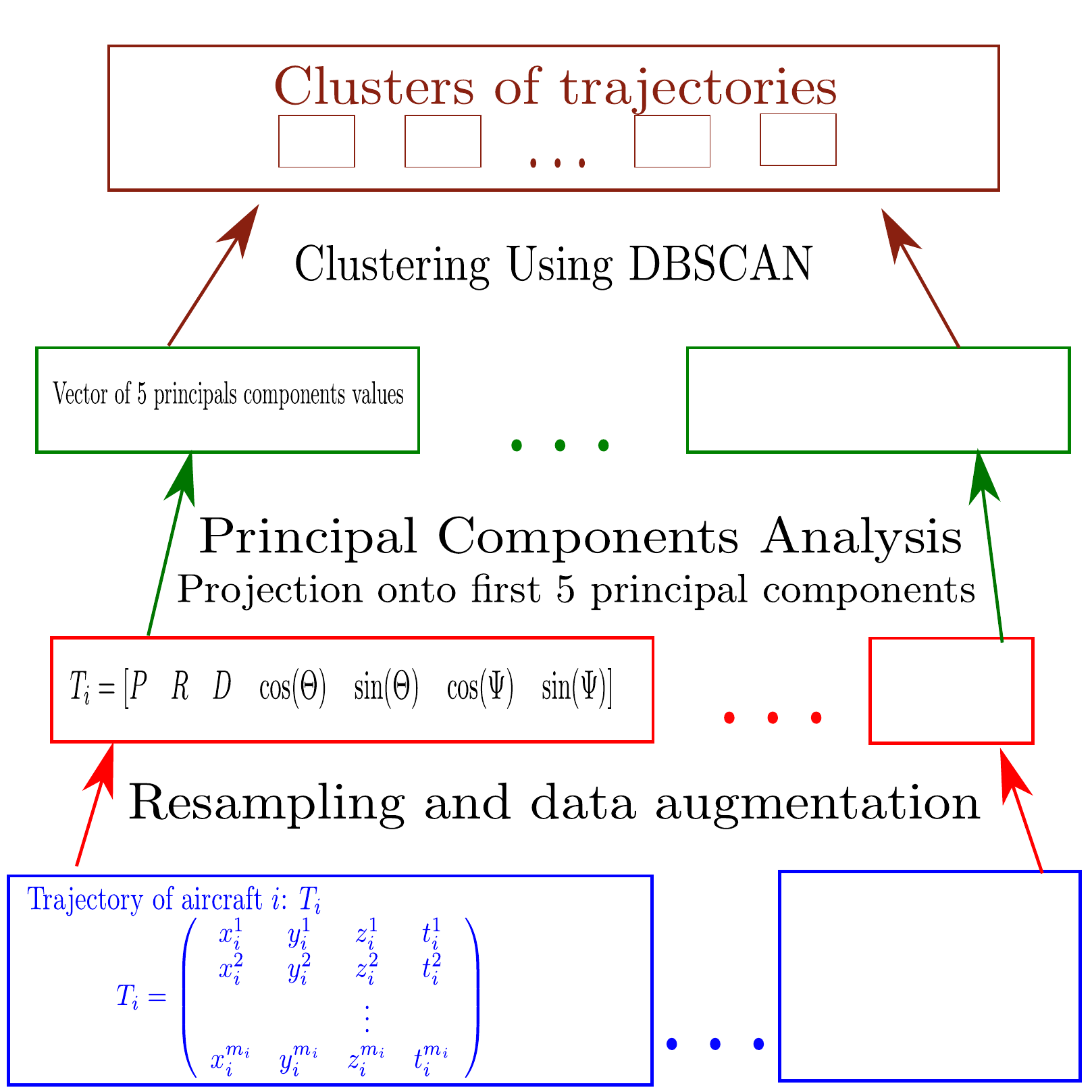}
\caption{Trajectory clustering method based on Principal Components Analysis}\label{fig:trajectoryClusteringPCAsummary}
\end{figure}

This  method proceeds with the following steps, which are summarized in a diagram in figure \ref{fig:trajectoryClusteringPCAsummary}:
\begin{enumerate}
\item Resample the trajectories, to obtain time series of equal length for each aircraft.
\item Augment the dimensionality of the data.
\item Normalize and concatenate all the data into a single vector for each flight.
\item Run a principal components analysis (PCA) and keep the first 5 principal components (PCs).
\item Cluster using a density-based clustering algorithm.
\end{enumerate}

\begin{figure}
\centering
 \subfigure[Clusters of trajectories ]{\includegraphics[width = 0.45\textwidth]{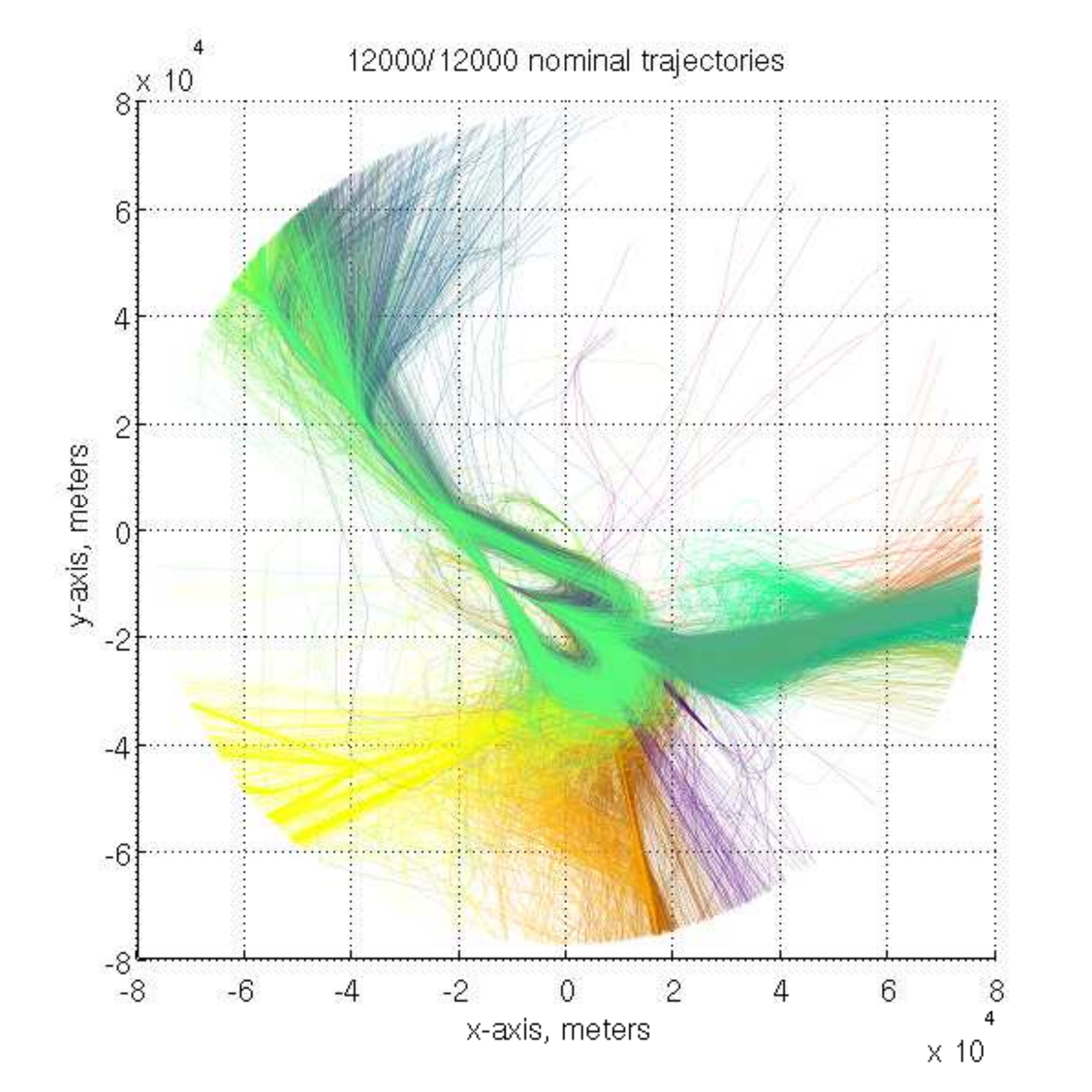}\label{fig:trajectoriesKmeansClustering2PC}}
\subfigure[Clusters of the first principal components]{\includegraphics[width = 0.45\textwidth]{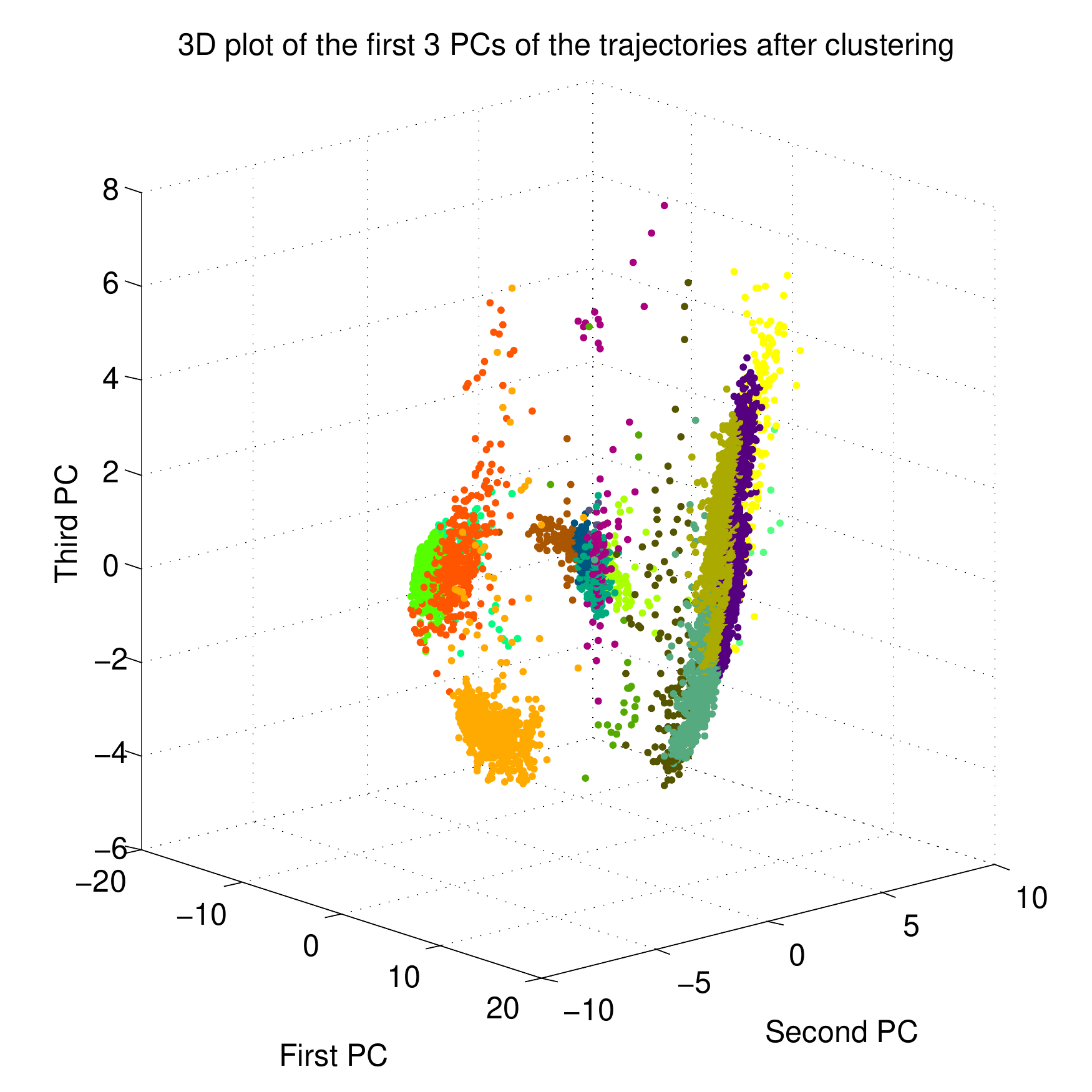}\label{fig:PCAclustersKmeans}}
\caption{Clustering results using the method presented in~\cite{eckstein09taxonomy}}\label{fig:kmeansClustering2PC}
\end{figure}

This paper proposes improvements to the approach used by Eckstein~\cite{eckstein09taxonomy} to realize a trajectory taxonomy. In~\cite{eckstein09taxonomy}, trajectories are first resampled, then the principal components are extracted and finally, the clustering is realized using $k$-means on the projections onto the first two principal components. Figure \ref{fig:trajectoriesKmeansClustering2PC} presents the resulting clusters on the principal components and on the trajectories, using the methods introduced in~\cite{eckstein09taxonomy}. The clustering technique proposed in~\cite{eckstein09taxonomy} does not provide result precise enough for our data set and there is no identification of outliers. Figure \ref{fig:PCAclustersKmeans} presents a 3D view of the projection onto the first three PCs (section \ref{sec:PCAs}) so it can be compared with our method. Eckstein used only the first two PCs for clustering. 
The first improvement is to augment the dimensionality of the data. Then, the PCs are computed and the projections of the augmented trajectories onto the first five PCs are clustered using a density based clustering algorithm. This algorithm presents the advantage of identifying outliers. Another advantage is that the number of clusters is not set a priori.

\subsubsection{Data Cleaning and Formatting}
The dataset is well organized and fairly clean. Trajectories with fewer than 50 points are removed from the dataset: to be able to use a  clustering algorithm such as DBSCAN, each trajectory must be represented as a vector. All vectors must have the same number of elements $n$, so their distance can be computed. Since all trajectories do not have the same number of points, resampling is necessary. Trajectories are resampled so that the total number of points for each trajectory is 50. For the sole purpose of clustering, fewer than 50 points would have been enough. Nevertheless, to improve the accuracy of the airspace monitoring function presented in section \ref{sec:airspaceMonitoring}, 50 points were used. The resampled trajectory $T'_i$ is given by ${T^{samp}_i = \big\{ T_i^l, l = \{\text{round}(\frac{k~m_i}{50}), k=1 \ldots 50\}\big\}}$. During this operation, the notion of speed that was given by the distance between the radar echos is lost.

\subsubsection{Dimensionality augmentation}
To improve the results of the clustering, the dimensionality of the data was increased. Some of the added dimensions present symmetry with respect to a point or a line and some do not.
\begin{itemize}
\item  Cartesian position of the aircraft in the resampled trajectory: ${P = [x_i^1\ldots x_i^{50} y_i^1\ldots y_i^{50} z_i^1\ldots z_i^{50}]}$. $P$ is a row vector with 150 components. This vector is unique to each trajectory.  		
\item 	Distance from the center of the TRACON ${R = \{r_i^l = \sqrt{(x_i^l)^2 + (y_i^l)^2 + (z_i^l)^2}, l = 1\ldots 50}\}$. Provides information about the rate of convergence of the aircraft toward the center of the TRACON, which is located close to the airport. This distance presents a symmetry with respect to the center of the TRACON, i.e two trajectories that are symmetric with respect to the center of the tracon will be represented with the same vector $R$.  
\item 	Distance from the top left corner: $D=  \{d^l_i=\sqrt{(x_i^l-x_{ref})^2 + (y_i^l-y_{ref})^2 + (z_i^l)^2}, l=1\ldots50\}$, where $(x_{ref}, y_{ref})$ are the coordinates of the top left corner a square containing the TRACON. The top left corner has coordinates $(x_{ref}, y_{ref}) = (-80, 80) km$. This distance presents a symmetry with respect of the diagonal joining the top left corner $(-80, 80)$ and the bottom right corner ($(80, -80))$, i.e two trajectories that are symmetric with respect to this diagonal will be represented with the same vector $D$.
\item 	Angular position in cylindrical coordinates: ${\Theta = \{\theta_i^l = \arctan(\tfrac{y_i^l}{x_i^l}), l=1\ldots50\}}$. A constant value and slow rate of change indicate a straight trajectory while high variablity indicates curved trajectories. Does not present any symmetry.	 \item 	Heading of the aircraft ${\Psi = \{\psi_i^l, l=1\ldots50\}}$. The computation of the heading was done using the filter of equation \ref{eq:headingFiltering} and then resampled to 50 points. This vector is unique to each trajectory.		 
\end{itemize}	
The sine and cosine values of the angular position and heading are used instead of their actual value to avoid the discontinuity at $\pm \pi$.
Each trajectory is now represented by a vector of dimension 450 given by: $T^{augm}_i = {[P \quad R \quad D \quad \cos(\Theta) \quad \sin(\Theta)  \quad \cos(\Psi) \quad \sin(\Psi)]}$. The initial vector had dimension 150.
The values of each parameter are normalized between 0 and 1 in order to balance their importance during the clustering process. It was decided to add meaningful data such as heading or rate of convergence toward the center of the TRACON. For instance, two aircraft on parallel trajectories will fly the same heading, even if the trajectories are slightly apart from each other. Distance to the center was chosen to identify trajectories that have particular patterns such as vectoring and holding pattern: the distance to the center will present some irregularities as the aircraft flies back and forth. Such irregularities will be highlighted by dimensions such as the heading that will change $180\dg$ while the position will only change slightly. 

\subsubsection{Principal Components Analysis}\label{sec:PCAs}
A principal components analysis~\cite{shlensTutorialPCA} is run on matrix that contains all the resampled trajectories. 
Each trajectory is then projected onto the first $p$ principal components and is now represented by a vector of $p$ values. The choice in the value of $p$ is a trade-off between computational speed when $p$ is small and accuracy when $p$ gets larger. There is no need to get a value of $p$ too large since the first principal components contain most of the information. Different values of $p$ were tried and $p = 5$ gave a satisfactory level of accuracy for this type of data.   
The added dimensions increse the range  of the projection of the trajectories onto the principal components. This makes the clustering task easier as the clusters are ``further apart'' in the principal components space. 

\subsubsection{Clustering}\label{sec:clusteringTraj}
\begin{figure}[ht]
\centering
 \subfigure[Clusters of trajectories ]{\includegraphics[width = 0.45\textwidth]{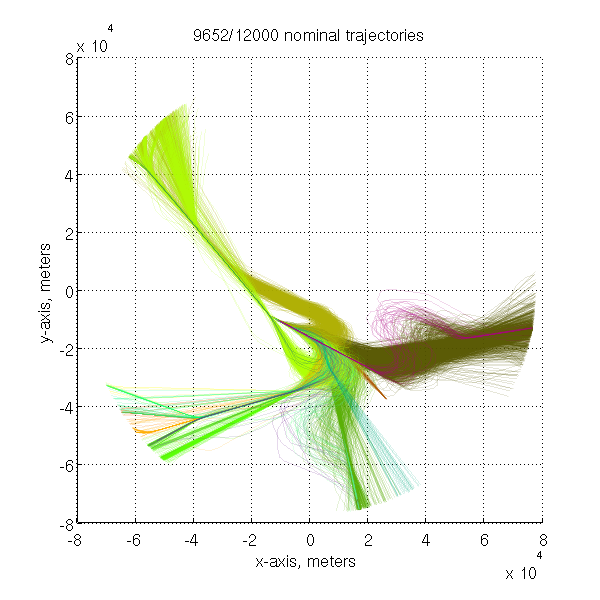}\label{fig:PCAtrajectoriesClusters}}
\subfigure[Clusters of the first 5 principal components]{\includegraphics[width = 0.45\textwidth]{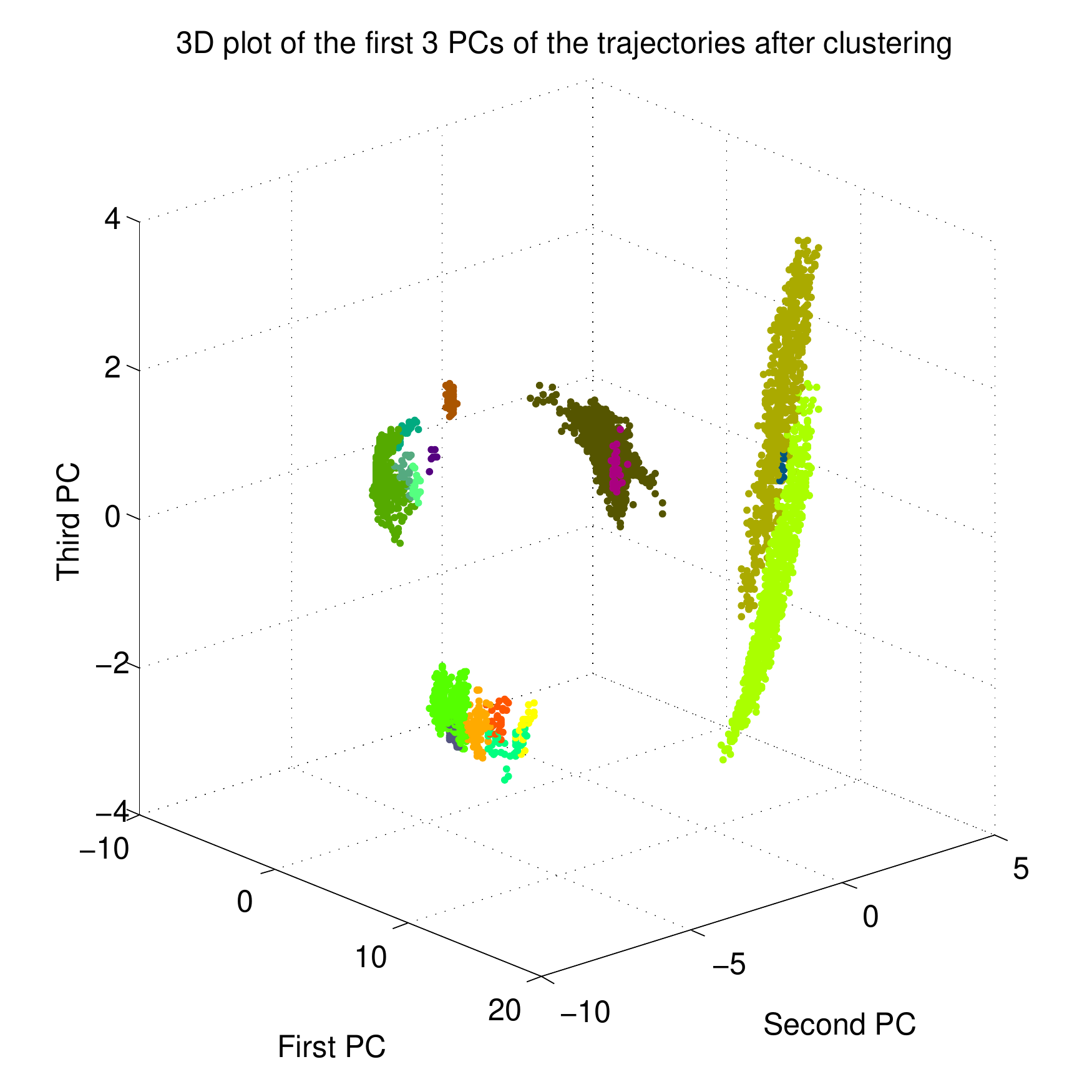}\label{fig:PCAclustersDBscan}}
\caption{Clustering results using resampling, data augmentation, PCA decompostion, and DBSCAN on the first 5 principal components.}\label{fig:DBscanClusterResults}
\end{figure}

The projections of the trajectories onto the first 5 PCs are clustered using DBSCAN. A density based clustering algorithm like DBSCAN is prefered to a distance based algorithm because of the shape of the clusters that can be arbitrary. The other advantage of DBSCAN is the identification of outliers. Figure \ref{fig:PCAclustersDBscan} presents the resulting clusters. The axis corresponds to the value of the first 3 principal components. Clusters are clearly differentiated, even if they are not easy to distinguish on the plot due to the perspective effect. The resulting clusters of trajectories are visually very clean (Figure \ref{fig:PCAtrajectoriesClusters}). 

\begin{figure}[ht]
\centering
 \includegraphics[width=0.45\textwidth]{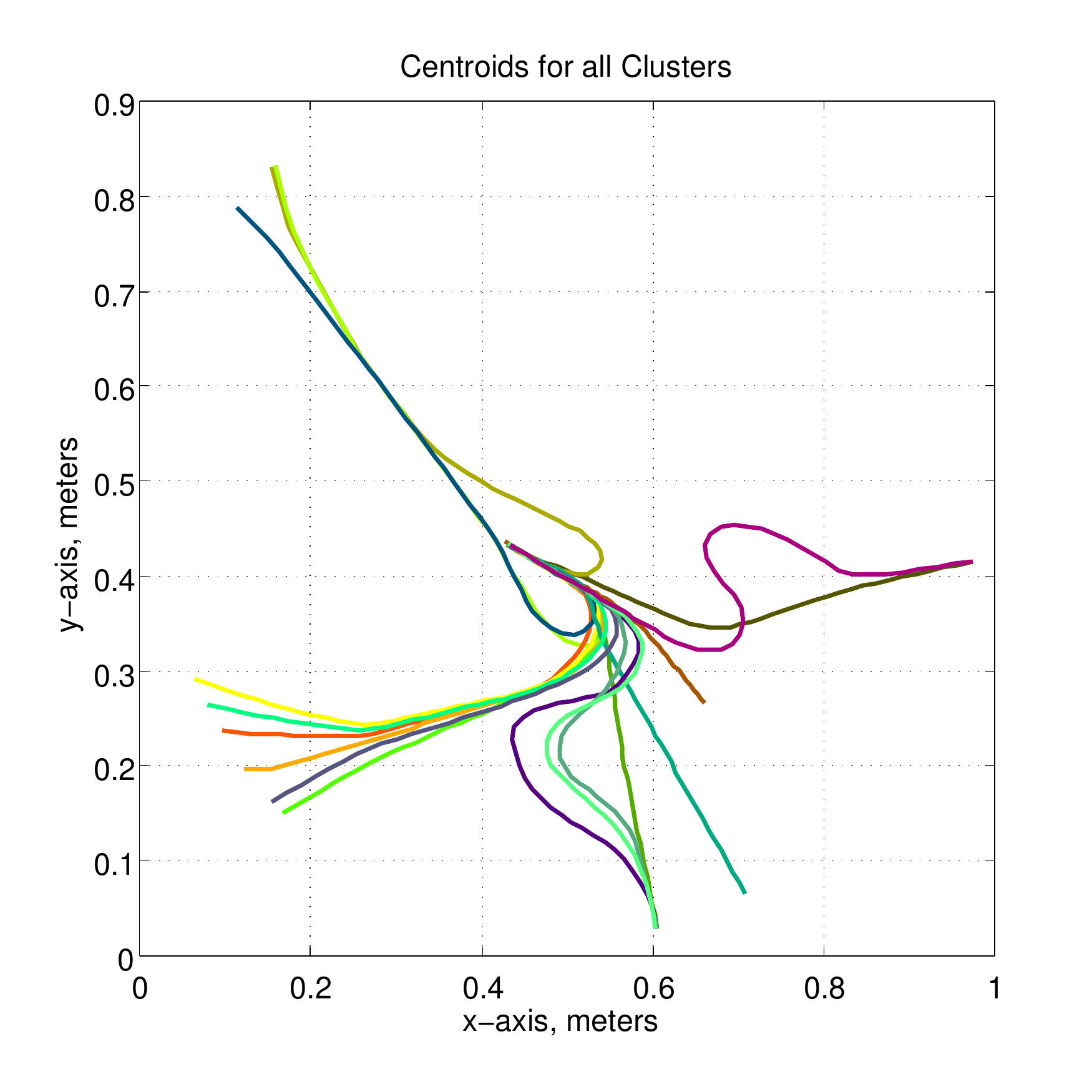}
\caption{Clusters centroids (average trajectory)}\label{fig:centroidsDBscan}
\end{figure}

Figure \ref{fig:centroidsDBscan} presents the centroids, that is the center of mass of the trajectories of each cluster. Those centroids can be seen as ``standard procedures''. Some clusters are minor variations from each other, such as the flights coming from the bottom left corner. This comes from the the settings used for DBSCAN. On Figure \ref{fig:PCAclustersDBscan}, one can clearly identy clusters of points. The algorithm was run with a high sensitivity ($\epsilon$ small and $minPts$ large). The parameter $\epsilon$ reflects the similarity between trajectories (the smaller the more similar), and $minPts$ is the number of ``similar'' trajectories needed to create a new cluster (Section \ref{sec:DBSCAN}). A small $\epsilon$ generates ``narrow'' cluster while a larger $\epsilon$ will generate clusters with more variability in trajectories. In the application presented in section \ref{sec:airspaceMonitoring}, the algorithm is run with a lower sensitivity and provides fewer clusters, with larger variability. The resulting centroids of this run can be seen on Figure \ref{fig:monitoring}.

\subsubsection{Analysis of outliers}

\begin{figure}[ht]
\centering
 \includegraphics[width=0.45\textwidth]{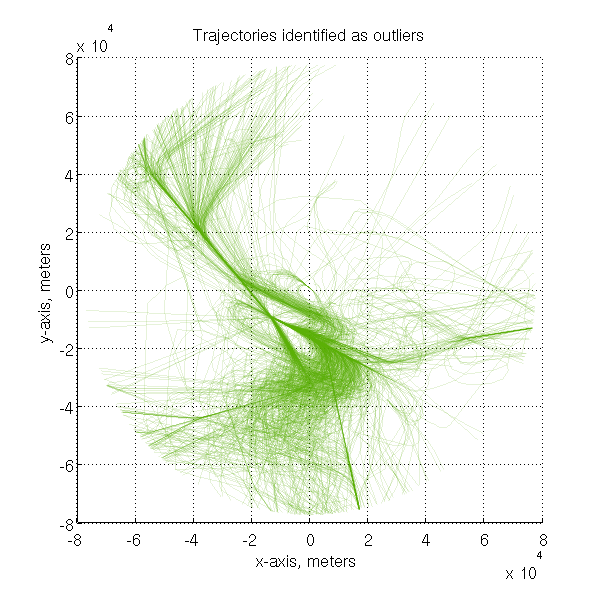}
\caption{Trajectories identified as outliers}\label{fig:outliersDBscan}
\end{figure}

Figure \ref{fig:outliersDBscan} shows the outliers detected by the clustering algorithm. Outliers represent 19.5\% of all trajectories.  A visual inspection shows that the main reasons for being detected as an outlier is the presence of holding patterns, large vectoring maneuvers or direct routes. 

 \begin{figure}[ht]
\centering
 \includegraphics[width=0.45\textwidth]{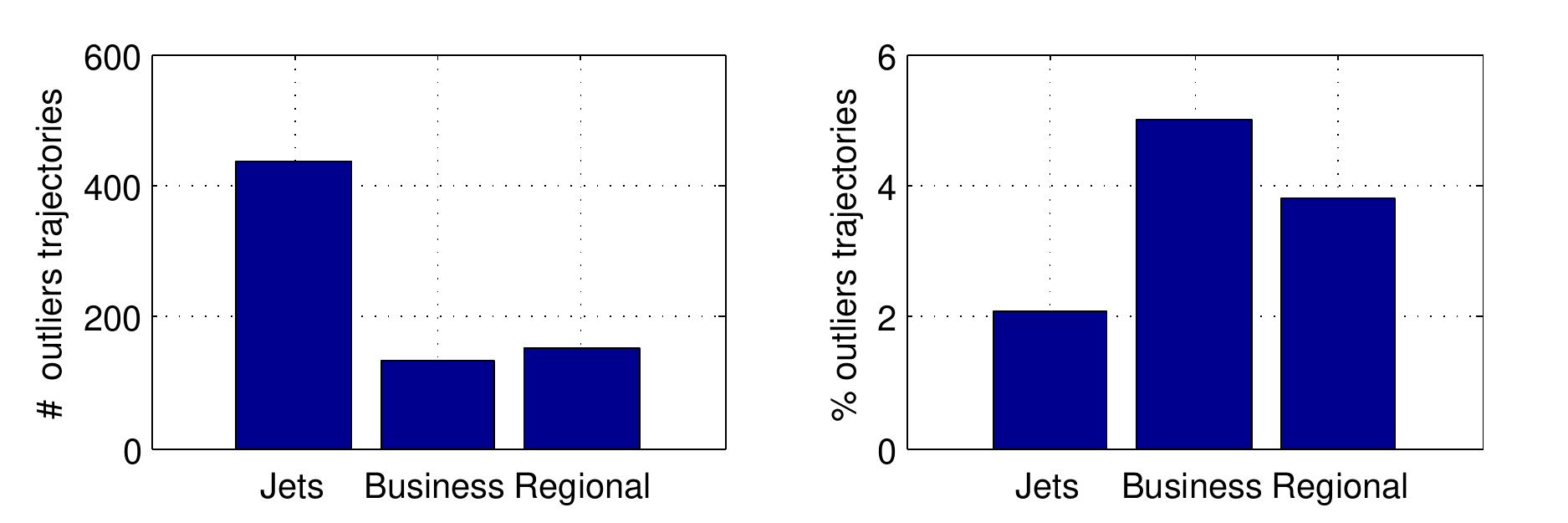}
\caption{Repartition of outliers by aircraft category}\label{fig:outliersAircraftType}
\end{figure}

Figure \ref{fig:outliersAircraftType} presents the number and frequency of outlier trajectories as a function of the type of aircraft. Jets represent the largest share in numbers, but the frequency is much smaller. Among the trajectories of regional and business aircraft, 4\% and 5\% are identified as outliers, respectively. A possible explanation is the size, the speed and the maneuverability of the aircraft. To ensure a safe separation at the runway threshold, air traffic controllers ``vector'' aircraft, that is give a sequence of headings to follow. The vectors given to business and regional aircraft might be different and sharper than the vectors given to larger size jets.

\begin{figure}[ht]
\centering
 \includegraphics[width=0.45\textwidth]{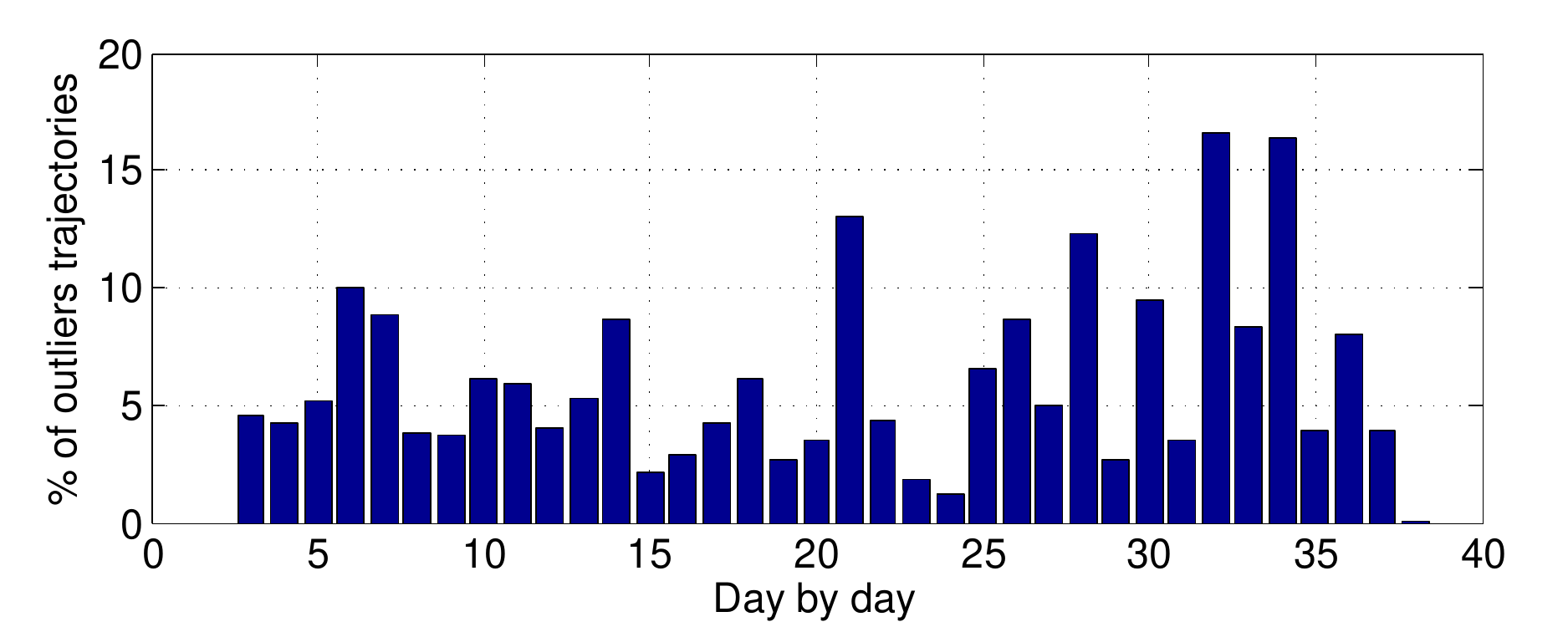}
\caption{Histogram of outliers, day by day }\label{fig:outliersDayOfTheMonth}
\end{figure}

Figure \ref{fig:outliersDayOfTheMonth} presents the frequency of outliers for each day of study. Each bar represents one day. The minimum percentage of outliers is less than 2\% and goes up to 16\%. The most likely explanation is the weather. San Francisco airport usually operates with two close  parallel runways. The runways are not independent, that is, they cannot be operated simultaneously when the weather does not permit visual approaches. When a runway is closed, the landing capacity is reduced from 60 to 30 aircraft per hour. Schedules and operations usually take the weather into account , but unexpected late fog dissipation or other type of convective weather might disrupt the operations and force controllers to vector aircraft and put them on holding patterns.

\begin{figure}[ht]
\centering
 \includegraphics[width=0.45\textwidth]{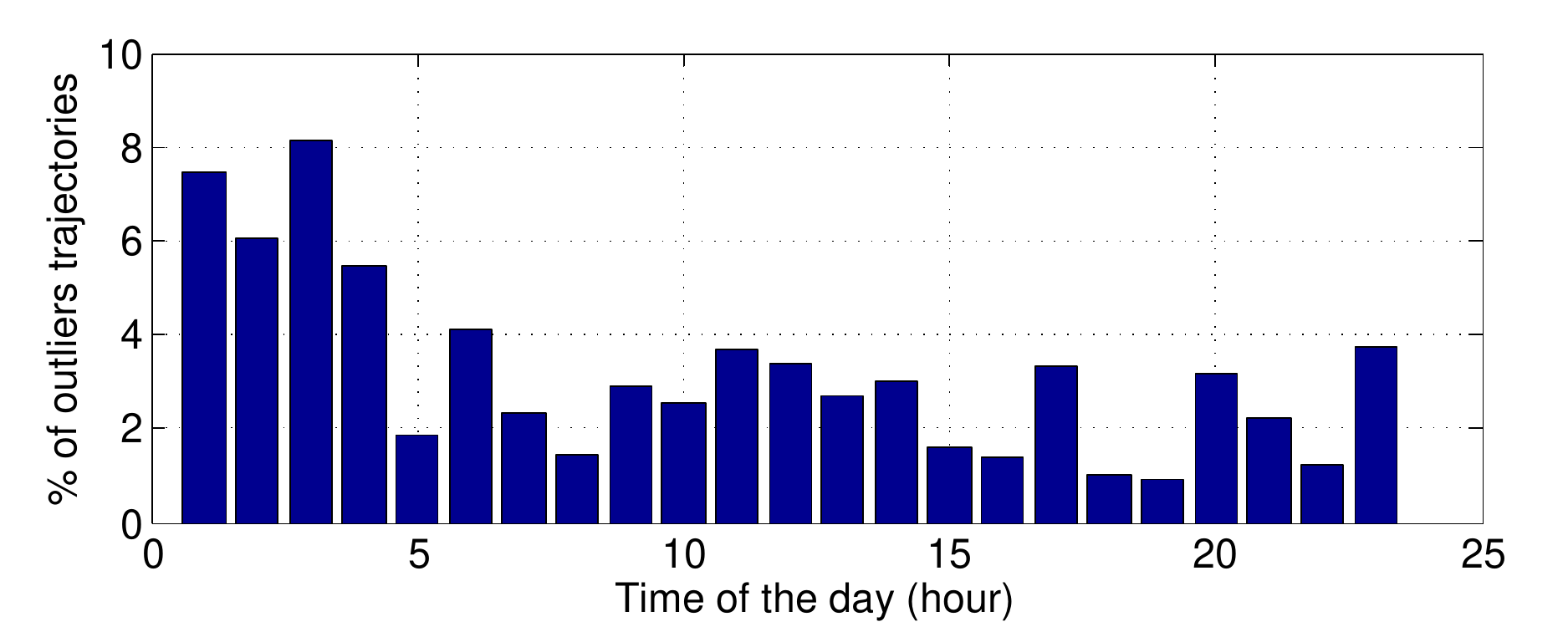}
\caption{Histogram of outliers, hour by hour, local time }\label{fig:outliersTimeOfTheDay}
\end{figure}

Figure \ref{fig:outliersTimeOfTheDay} presents the frequency of outliers as a function of the time of the day. The local time is reported on the abscissa axis, starting at midnight. This diagram is an average over the entire period of interest. The frequency of outliers is higher during the night, then decreases in the early morning, to in increase again with a peak at 11a.m.. Another peak is visible at 5 p.m.. The outliers identified at night are mostly due to direct routing that is allowed by the very low traffic density at night. During the morning, traffic density increases and requires more rerouting for efficient sequencing and merging. Another possible explanation is the late dissipation of the fog.   


\section{Airspace monitoring}\label{sec:airspaceMonitoring}
This section proposes an airspace monitoring technique that automatically detects when an aircraft is not conforming to standard procedures. Standard procedures are determined using the centroids of the clusters found using the previous clustering methods. Centroids correspond to flight path often flown and the value of the parameters used for clustering allows the trajectories to vary more around the centroids.  The objective of the monitoring task is to detect in  when an aircraft deviates from nominal path in real time.

\subsection{Literature review and motivation}
Krozel~\cite{krozel2002intelligent} proposed an intent based monitoring where the aircraft is tracked relative to a filed flight plan, using NavAids and waypoints. The monitoring tasks requires knowledge of the airspace structure, of the trajectory waypoints and of the intent of the aircraft. It is a powerful tool when the flights behave according to their flight plan, but when dealing with arrivals, the sequence of waypoints might change, some might be skipped or added to ensure an optimal separation of aircraft at the runway threshold and vectoring is often used. This monitoring method cannot be used. 
 Reynolds et al.~\cite{reynolds2002structure} introduced a framework for the development of an automated conformance monitoring system. The system described in~\cite{reynolds2002structure} has two main inputs: the conformance basis, containing target states and trajectory information, and the observation of a surveillance system. Those inputs feed models for pilots intents, aircraft intents, and aircraft control systems and dynamics. Those models provide an expected state vector that is compared with the observed state vector for conformance analysis. This structure is further used in~\cite{reynolds2002conformance} to monitor the conformance of a trajectory to a flight plan. For instance, it detects if an aircraft does not turn, turns too early or too late at a waypoint. The monitoring is based on intent and knowledge of the the exact expected trajectory is required. An offline trajectory analysis and taxonomy for arrival trajectories was proposed by Eckstein~\cite{eckstein09taxonomy}. The objective of Eckstein is to analyze the performance of area navigation (RNAV) operations for NextGen concepts of operations offline. The method in ~\cite{eckstein09taxonomy} uses GPS coordinates of actual waypoints to identify and classify segments of trajectories. 

\begin{figure}[ht]
 \centering
\includegraphics[width = 0.45\textwidth]{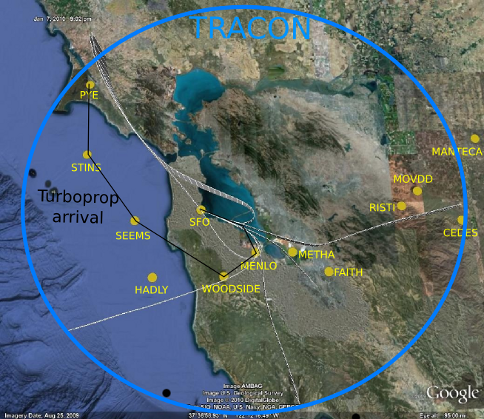}
\caption{Centroids of the clusters and reporting points/waypoints for SFO arrivals}\label{fig:arrivalsWaiPointsSFO}
\end{figure}

Figure \ref{fig:arrivalsWaiPointsSFO} displays an aerial view of the San Francisco Bay Area. The blue circle represents the outer boundary of the TRACON, given by the area covered by the radar. The white lines are the centroids identified in section \ref{sec:clusteringTraj}. The yellow dots are waypoints or reporting points. The locations of those points come from Standard Terminal Arrival Routes (STAR) and track logs \cite{flightAware}. The centroids of the clusters pass over only a limited number of waypoints. This shows that using the published waypoints and reporting points cannot be efficiently used to monitor traffic in the TRACON. The intent based methods cannot be used in the terminal area.  Figure  \ref{fig:arrivalsWaiPointsSFO} also displays the arrival from the north for turboprops. This arrival procedure has not been identified by the clustering algorithm because of the relatively small number of aircraft using this route, and of the variability of the flight path following this procedure.  
A real time trajectory analysis tool built upon the knowledge gathered from the clustering analysis is now proposed. The tool is called AirTrajectoryMiner (ATM) since it enables the monitoring of operations in the TRACON. Current aircraft trajectories are compared against nominal trajectories, that is the trajectories in the clusters. If they differ too much, the current trajectory is tagged as abnormal, or outlier. The only intent used is the aircraft final destination airport. The tool automatically detects if the aircraft is flying one of the possible approaches, including most commonly used vectoring maneuvers.

\subsection{Data formatting}
It is not possible to directly compare the current trajectories with nominal trajectories, since current trajectories are incomplete. During real-time system operations, only past data are known. Therefore, the nominal dataset is fragmented. Resampled trajectories that had 50 points are split in 10 fragments of 5 points. The average travel time in the TRACON for aircraft landing at SFO is about 14 minutes. Therefore, 5 data points correspond to about 14*60/10 = 84 seconds. A memory of 80 seconds is used for current tracks. The radars hits of the last 80 seconds of flight are resampled to 5 points that can now be compared against the database of nominal tracks. This process is done using the Inductive Monitoring System.  

\subsection{Anomaly detection: Inductive Monitoring System}
To detect anomalous trajectories, the Inductive Monitoring Health System (IMS)~\cite{iverson2004inductive} is used. IMS  is a good alternative to model based health monitoring systems. It provides a high fidelity detection tool, and there is no need to manually build a model. 
IMS runs in two steps: learning phase and anomaly detection. IMS learns the nominal behaviors using a training dataset provided by the user. IMS builds clusters using $k$-means clustering and density-based clustering. During the anomaly detection phase, the input data is compared with knowledge base built from the training data. The anomaly score can be interpreted as the distance to the nearest cluster. The input data belongs to a cluster if all the parameteres values are within the range specified by the cluster limits. 

The training dataset corresponds all the trajectories identified as nominal and fragmented in 10 segments of 5 points. The total number of segments was 276,040.

\subsection{AirTrajectoryMiner: Monitoring tool}
\begin{figure}
 \centering
\includegraphics[width=0.48\textwidth]{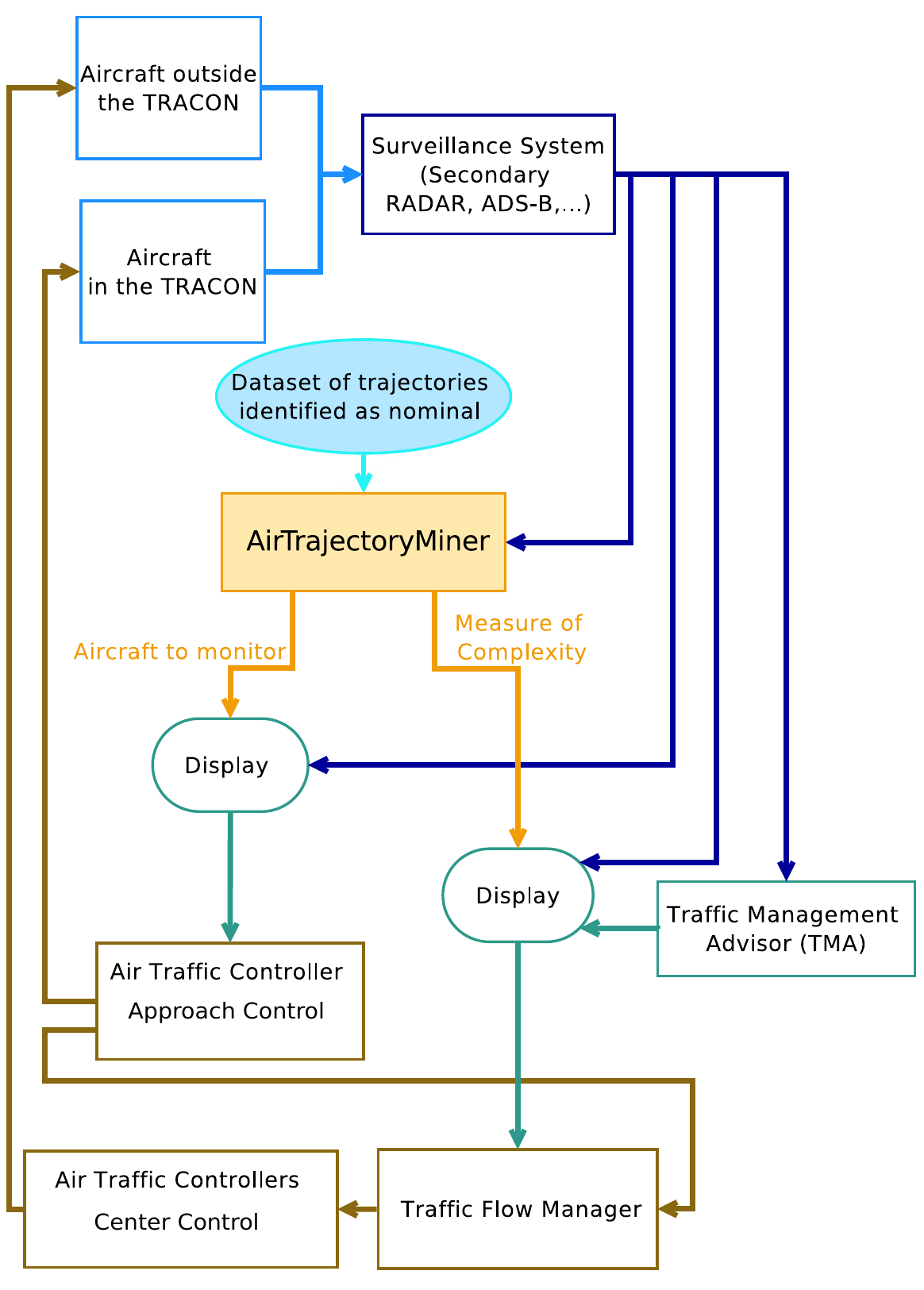}
\caption{Schematic view of the air traffic control system in and around the TRACON - Integration of AirTrajectoryMiner}\label{fig:diagramToolUsefulness}
\end{figure}

AirTrajectoryMiner (ATM) is a  real time TRACON monitoring tool. Figure \ref{fig:diagramToolUsefulness} shows how ATM could be incorporated into the air traffic management environment. The inputs to the tool are the set of all the trajectories identified as nominal, work resulting from section \ref{sec:clusteringTraj}, and the radar tracks of the flights of interest. ATM delivers two types of ouputs. On the one hand, it delivers an indication of conformance of current flight to nominal procedures, and on the other hand, it delivers a measure of the complexity in the TRACON that can be incorporated in Traffic Management Advisor (TMA) software\cite{lee2000human}. 

\begin{figure}[ht]
\centering
\includegraphics[width=0.45\textwidth]{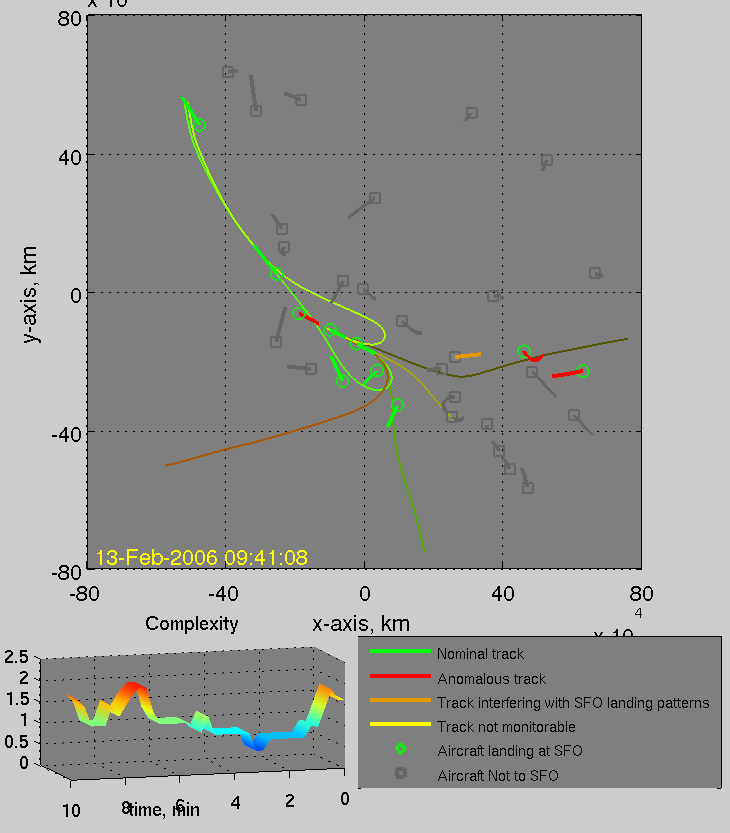}
\caption{AirTrajectoryMiner display. Top frame: conformance to standard procedure. Bottom frame: time history of complexity in the TRACON.}\label{fig:monitoring}
\end{figure}
Figure \ref{fig:monitoring} displays the monitoring environment. The top frame is a 2D view of the airspace. The records cover a cylinder of radius 80km, going from the ground up to 6000 m. 
The following provides some information about the display and associated aircraft count.
\begin{itemize}
 \item \textbf{Green circle}: aircraft intended to land at SFO (associated count: $n_{SFO}$).
 \item \textbf{Grey square}:  aircraft not intended to land at SFO (associated count: $n_{\overline{SFO}}$).
\item \textbf{Green segment}: trajectory of an aircraft intended to land at SFO and following the procedures.
\item \textbf{Red segment}: trajectory of an aircraft intended to land at SFO and whose trajectory is identified as an outlier: it does not to follow the procedures (associated count: $n_{\overline{OK},SFO}$).
\item \textbf{Grey segment}: trajectory of an aircraft not intended to land at SFO and whose trajectory does not interfere with traffic landing at SFO.
 \item \textbf{Orange segment}: trajectory of an aircraft not intended to land at SFO and whose trajectory may interfere with traffic intended to land at SFO (associated count: $n_{\overline{OK},\overline{SFO}}$).
\item \textbf{Colored lines}: Centroids of the clusters of trajectories identified as nominal. The centroids presented differ from the ones on Figure \ref{fig:centroidsDBscan}, because the clustering algorithm was re-run with different parameters allowing more variability and therefore creating fewer clusters.
\end{itemize}

 Aircraft intent information comes from the data. The length of the line following the aircraft corresponds to the part of the trajectory being analyzed, that is the last 80 seconds of the trajectory. The length of the line is therefore proportional to the velocity of the aircraft. This display is intended for an air traffic controller managing the arrivals at SFO. A similar display would be used for managing other arrivals or departures. The only change would be the training data for IMS and the centroids displayed. Aicraft with a grey segment can be ignored, since they are not landing at SFO and are not interfering with landing traffic at SFO. Aircraft with a green segment are following the standard procedures to land at SFO. Aircraft in orange require special attention since they are not intended to land at SFO but present characteristics that identify them as ``in the pattern to land at SFO''. They conform with some of the SFO landing trajectories. Aircraft in red also need special attention since they are supposed to land at SFO but currently not on standard tracks. The controller needs to make sure they are not generating conflicts or interfering with other traffic. Based on the compliance of current flights to procedures, we define a measure of complexity for the TRACON, which could provide an automatic feedback of the health of the TRACON to the traffic flow manager who regulates the flow of aircraft arriving in the TRACON.

\subsection{Measure of complexity}
Complexity in air traffic management is a widely studied topic~\cite{mogford1995cca}. A measure of airspace complexity is called dynamic density~\cite{airspaceComplexitySridhar} and was intended to understand the effect of changing airspace configuration and traffic controller workload. It is a function of the traffic density and the number of aircraft changing heading, speed or altitude, and, the separation between aircraft. 
This measure is a weighted sum of several parameters and the weights have been determined using human in the loop experiments. The model was fitted to the observations of controllers.  Delahaye and Puechmorel~\cite{delahaye} propose a measure of complexity based on the Lyapunov exponents of a time varying vector field that interpolates aircraft position and velocities. This intrinsic complexity measure reflects the stability of the traffic configuration. Lee~\cite{keumjinLeeThesis} propose complexity measure based on the response of the airspace to a disturbance. Disturbance can be an intruder aircraft in the airsapce and the corresponding measure of complexity is the deviation required by the other aircraft to solve all the conflicts. Gariel et Feron~\cite{GarielATMcomplexity2008} proposed complexity maps based on the degradation of communication, navigation and surveillance capacities. The degradation results in a required increase in separation distances creating new potential conflicts. The complexity measures the difficulty to steer the traffic from the nominal mode of operation with initial separation distances to degraded mode of operation with increased separation distances.

This paper introduces a new complexity metric, based on the compliance of aircraft to procedures identified as nominal. According to ~\cite{mogford1997mental,histon2002sca}, controllers build a mental model of nominal operations. The complexity of a traffic configuration perceived by the controllers increases when aircraft flight path do not follow this mental model. When operations are running as expected, the controller is more efficient and can deal with more aircraft. Thus, increasing the number of aircraft not following nominal procedure will reduce the maximum number of aircraft a controller can deal with simultaneously, reducing the capacity of the airspace. The proposed complexity measure is based on Shannon's theory of communication~\cite{shannon2001mathematical}.
Using the aircraft counts introduced earlier, the instantaneous probability of an aircraft inbound for SFO to be identified as nominal is 
\begin{equation}
 p(OK|SFO) = \frac{n_{OK,SFO}}{n_{SFO}}.
\end{equation}
For the aircraft inbound for SFO and identified as outliers, it is assumed that each outlier aircraft is unique and independent from other aircraft. At each instant, each outlier is considered a different from the other outliers, that is there are $n_{\overline{OK},SFO}$ types of outliers. Therefore, the probability of an aircraft to be a specific outlier is 
\begin{equation}
 p(\overline{OK_i}|SFO) = \frac{1}{n_{SFO}}, \quad i=1\dots n_{\overline{OK},SFO} .
\end{equation}

The entropy $I_{SFO}$ of the aircraft inbound to SFO is therefore

\begin{small}
\begin{equation}
\begin{split}
 I_{SFO} =&-p(OK|SFO)\log{p(OK|SFO)} \ldots\\
&-\sum_{i=1}^{n_{\overline{OK_i},SFO}}p(\overline{OK_i}|SFO) \log{p(\overline{OK_i}|SFO)}\\
 =& - \frac{n_{OK,SFO}}{n_{SFO}}\log\frac{n_{OK,SFO}}{n_{SFO}} - \frac{n_{\overline{OK},SFO}}{n_{SFO}} \log{\frac{1}{n_{SFO}}}.
\end{split}
\end{equation}
\end{small}

The same reasoning is used for aircraft not inbound to SFO:
\begin{small}
\begin{equation}
 I_{\overline{SFO}} = - \frac{n_{OK,\overline{SFO}}}{n_{\overline{SFO}}} \log\frac{n_{OK,\overline{SFO}}}{n_{\overline{SFO}}} - \tfrac{n_{\overline{OK},\overline{SFO}}}{n_{\overline{SFO}}} \log{\frac{1}{n_{\overline{SFO}}}}.
\end{equation}
\end{small}

The proposed measure of complexity $C$ is the sum of the entropy of aircraft inbound to SFO and the entropy of aircraft not inbound to SFO. 

\begin{equation}
C = I_{SFO} + I_{\overline{SFO}}.
\end{equation}

This complexity measure is an indication of the disorder with regard to nominal operations. If no aircraft is identified as an outlier, the complexity is 0. The complexity increases with the number of outliers detected, but also with the number of aircraft. 
The bottom left plot of Figure \ref{fig:monitoring} shows this measure of the complexity over the last 10 minutes. The plot is refreshed every 15 seconds. When the traffic flow manager sees that the complexity increases, 
ATM provides information about the operations in the TRACON. If the complexity gets high, the controller in charge of the TRACON is likely to have a high workload. Providing the traffic flow manager with this complexity measure can help him to manage the flow of arriving aircraft. A low complexity suggests that more aircraft can be allowed in the TRACON. Increasing complexity suggests that the TRACON controllers are subject to an important workload and that the aircraft arrival rate should be reduced.

An aircraft can be detected as ``off nominal track'' because the controller may require a large vectoring maneuver, meaning that the runways are congested. Another explanation for ``off nominal'' tracks can be rerouting due to weather.  
This tool can also be used as an automatic independent monitoring tool. Intent based tools~\cite{reynolds2002conformance} cannot be used in terminal areas since controllers give vectors that do not appear in the flight plan. Moreover, there are many turns and altitude changes that are left to the pilot to execute.

\section*{Conclusion}
This paper presented two trajectory clustering methods and an application to airspace monitoring. The monitoring tool compare the conformance of current flights to identified nominal procedures in real-time. The version of the tool presented in this paper monitors the landings at SFO, but it can easily be modified to monitor any traffic pattern, by modifying the input dataset. 

\section*{Acknoldegments}
This work was supported by NASA under Grant NNX08AY52A and by Thales ATM inc.
\bibliographystyle{unsrt}
\bibliography{bibli}
\end{document}